\documentclass[10pt,twocolumn,letterpaper]{article}

\usepackage[pagenumbers]{cvpr} 
\usepackage[utf8]{inputenc} 
\usepackage[T1]{fontenc}    
\usepackage{booktabs}
\usepackage{multirow}
\usepackage{graphicx}
\usepackage{xcolor}
\usepackage{colortbl}
\usepackage{tcolorbox}
\tcbuselibrary{skins,breakable}
\definecolor{myPurple}{RGB}{105,76,145}

\definecolor{BenchPeach}{RGB}{251,240,230}
\definecolor{BenchBlue}{RGB}{231,242,250}
\definecolor{BenchLavender}{RGB}{238,235,250}

\newtcolorbox{promptbox}[2][]{
  enhanced,
  breakable,
  colback=white,
  colframe=myPurple,
  coltitle=white,
  colbacktitle=myPurple,
  fonttitle=\bfseries,
  title={#2},
  boxed title style={
    colback=myPurple,
    colframe=myPurple,
    rounded corners
  },
  attach boxed title to top left={
    xshift=0.6cm,
    yshift=-2mm
  },
  top=5mm,
  left=4mm,
  right=4mm,
  bottom=3mm,
  arc=2mm,
  boxrule=1.2pt,
  #1
}

\definecolor{cvprblue}{rgb}{0.21,0.49,0.74}
\usepackage[pagebackref,breaklinks,colorlinks,allcolors=cvprblue]{hyperref}

\def\confName{CVPR}
\def\confYear{2027}

\DeclareRobustCommand{\titleicon}{%
  \raisebox{-0.2\height}{%
    \includegraphics[height=1.9em]{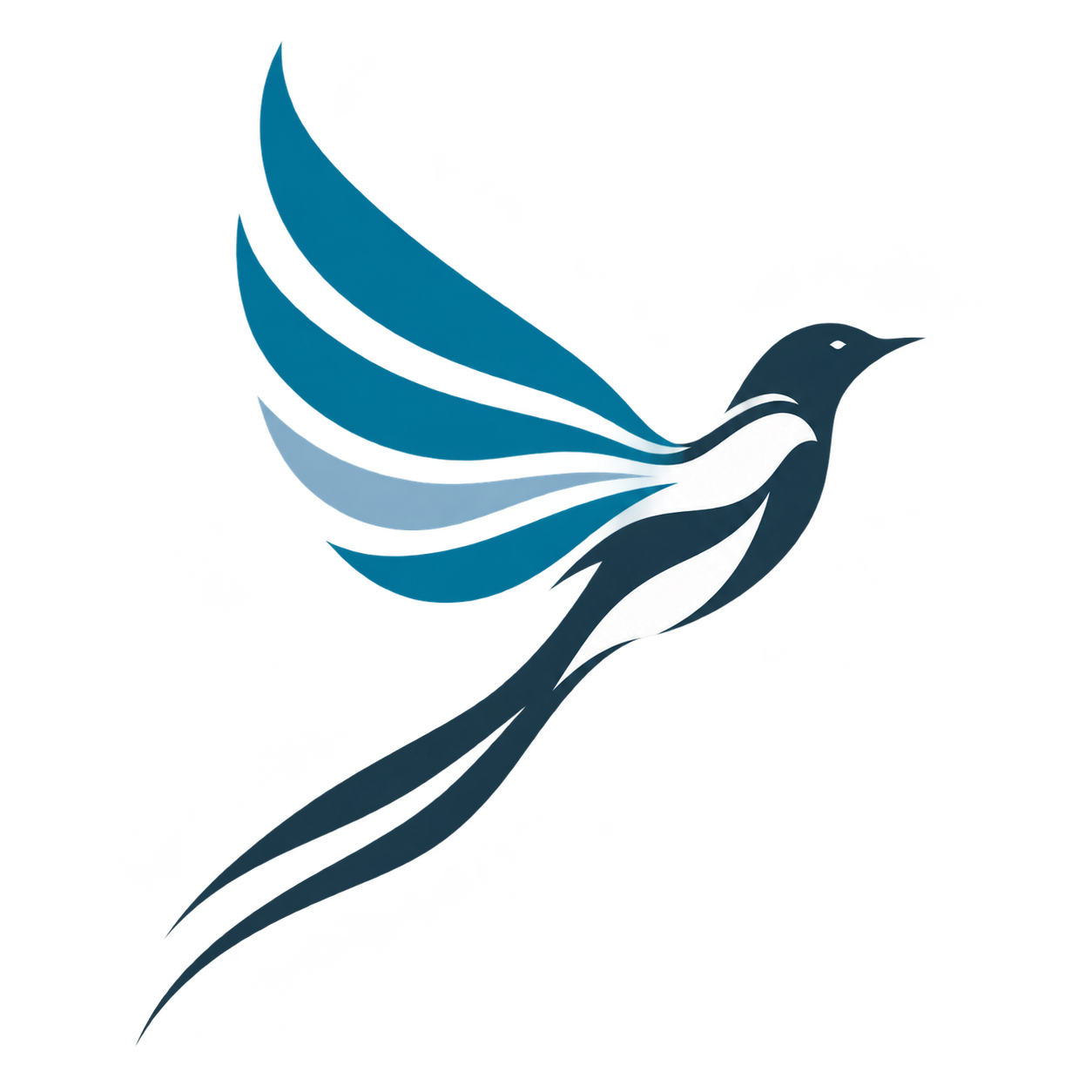}%
  }%
}

\title{%
  \texorpdfstring{%
    \titleicon\hspace{0.2em}%
    MAGPIE: Agentic Multi-Reference Image Generation
    via Executable Scene Construction%
  }{%
    MAGPIE: Agentic Multi-Reference Image Generation
    via Executable Scene Construction%
  }%
}

\author{
\textbf{Yifan Dai}$^{1}$\thanks{These authors contributed equally to this work.},
\textbf{Daili Hua}$^{2}$\footnotemark[1],
\textbf{Bohan Zeng}$^2$,
\textbf{Yuran Wang}$^{2}$,
\textbf{Jialing Liu}$^4$,
\textbf{Bozhou Li}$^{3}$,\\
\textbf{Chengzhuo Tong}$^{2}$,
\textbf{Yang Shi}$^{2}$,
\textbf{Kai Zeng}$^2$,
\textbf{Liu Yang}$^5$,
\textbf{Xinyu Liu}$^6$,
\textbf{Shujie Li}$^7$, \\
\textbf{Yiyan Ji}$^{8}$,
\textbf{Bingyin Mei}$^9$,
\textbf{Yushuo Guan}$^3$,
\textbf{Yuanxing Zhang}$^3$,
\textbf{Fangcheng Fu}$^1$,
\textbf{Wentao Zhang}$^{2}$\\
$^1$School of Artificial Intelligence, Shanghai Jiao Tong University,
$^2$Peking University,\\
$^3$KlingAI Research,
$^4$Renmin University of China,
$^5$SJTU,
$^6$HKUST,
$^7$HKU,
$^8$NJU,
$^{9}$THU \\
}

\begin{document}
\maketitle
\begin{abstract}
Multi-reference image generation requires a model to preserve the appearance of selected subjects while arranging them according to an instruction. Despite recent progress in reference-based generation and compositional control, reliably organizing selected subjects into the intended scene remains challenging. Although existing methods provide various forms of compositional guidance, the planned relationships are not always explicitly instantiated for inspection and refinement before image synthesis. To address this limitation, we introduce MAGPIE, which represents the intended composition as a simplified 3D scene defined by an executable Blender program. A coding agent links each scene instance to its reference, checks the geometry and rendered view against the instruction, and revises the program to correct discrepancies. The final proxy rendering serves as a visual plan for the image generator. Using MAGPIE-Instruct-3K, we train the generator to follow this plan while drawing each subject's appearance from its reference. MAGPIE improves both Qwen and FLUX on DreamOmni2 and SconeEval. With FLUX, it improves overall success on DreamOmni2 by 28.21 percentage points over the original backbone and achieves the highest overall score among the models evaluated on SconeEval.
\end{abstract}    
\section{Introduction}
\label{sec:intro}



\begin{figure}[!t]
  \centering
   \includegraphics[width=\linewidth]{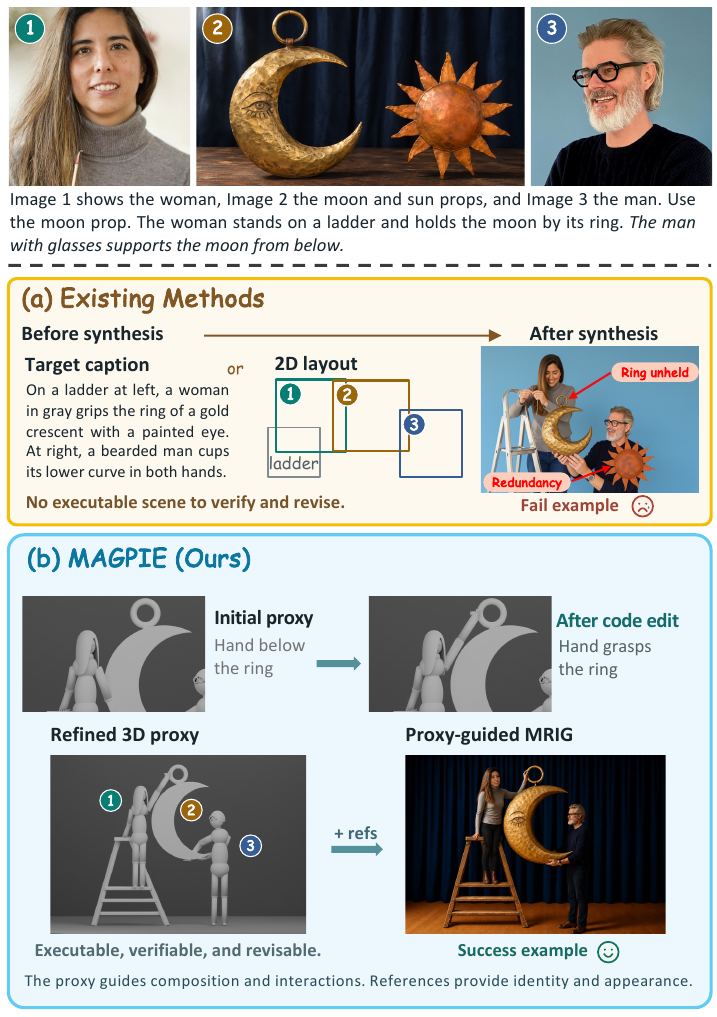}
   \caption{\textbf{Comparison of representative existing methods and MAGPIE.} (a) Existing methods provide semantic or spatial guidance without necessarily exposing an explicit scene for verification. (b) In contrast, MAGPIE instantiates compositional intent as an executable 3D proxy, enabling revision before reference-grounded appearance synthesis.}
   \label{fig1}
\end{figure}

Multi-reference image generation (MRIG) enables users to compose personalized images from heterogeneous visual references and natural language instructions~\cite{wu2025qwen, Xiao_2025_CVPR,flux-2-2025,Wang_2026_CVPR,Xia_2026_CVPR,wu2026omnigen2}. Recent advances have improved reference fidelity and instruction understanding: DreamOmni2~\cite{Xia_2026_CVPR} supports multimodal generation and editing, while Scone~\cite{Wang_2026_CVPR} explicitly addresses subject selection among multiple candidates. However, faithfully preserving the appearance of referenced subjects does not necessarily ensure that they are composed correctly~\cite{Wang_2026_CVPR,Oshima_2026_CVPR}. As instructions require selective inclusion, multiple instances, and interactions among subjects, the generator must organize them into a scene that satisfies the requested spatial and interaction relationships.

This difficulty motivates a closer look at how direct-generation pipelines organize a scene~\cite{wu2026omnigen2,Xia_2026_CVPR,Xiao_2025_CVPR}, as illustrated in Fig.~\ref{fig1}(a). Within this paradigm, a generative model must implicitly determine which entities should appear and how they should be organized while simultaneously synthesizing their reference-specific appearances. Without an explicit intermediate scene, compositional errors such as a missing instance or an incorrect spatial relationship become apparent only in the generated image.

Recent approaches introduce semantic or spatial guidance through textual reasoning and structural conditions~\cite{He_2026_CVPR,feng2023layoutgpt,ICLR2026_9b01b794,Shi_2026_CVPR,team2026behind}.
However, specifying the intended composition does not necessarily instantiate its fine-grained geometry. As shown in Fig.~\ref{fig1}(a), a plan may specify a woman holding a ring without resolving the hand--ring contact. The same bounding boxes allow different poses and depth orderings. Leaving these choices to the generator makes it difficult to verify the requested relationships or revise the geometry before synthesis. Geometry-aware approaches further explore explicit 3D scene planning and agentic revision, enabling more direct control over spatial relationships~\cite{Ding_Zhuang_Li_Yue_Qiao_Wang_2025,qian-etal-2026-spatialguard}. While these methods primarily focus on text-driven image generation, multi-reference generation introduces an additional challenge: maintaining the correspondence between references and scene instances while transferring the planned composition into reference-faithful image synthesis.

Executable scene code lets us inspect spatial relationships in the resulting geometry and rendered views and revise individual placements and poses before image synthesis. Drawing on these properties, we introduce MAGPIE, a framework for multi-reference image generation through verifiable scene construction. MAGPIE constructs executable 3D proxies whose entities retain explicit bindings to the original visual references. As shown in Fig.~\ref{fig1}(b), the agent executes each scene program, checks the geometry and rendered views against the requested relationships, and iteratively revises the program to correct discrepancies. The refined proxy rendering then guides composition, while the bound references provide each subject's appearance for the final image generator.

Making verifiable scene construction useful for MRIG requires addressing two connected challenges. First, the agent must reliably construct and repair a proxy that captures the composition specified by the references and instruction. Second, the generator must learn to follow the proxy's structure while preserving appearance from the corresponding references. To address the first challenge, we develop a structured agentic harness that integrates scene planning, procedural Blender construction, and iterative geometric and visual revision. Parameterized components and shared interaction anchors provide controllable geometry for individual entities and their interactions. Because a program can execute successfully while placing entities incorrectly, the harness checks entity visibility, camera-relative ordering, and designated contacts on the constructed geometry. A multimodal reviewer complements these measurements by examining the rendering against the references and instruction, and the agent uses the combined feedback to revise the scene.

To address the second challenge, pretrained generators must learn to use simplified proxy renderings as structural guidance without copying their placeholder appearances. However, existing MRIG datasets generally lack supervision that aligns executable scene proxies with visual references and target images. We therefore construct MAGPIE-Instruct-3K, a dataset of 3187 examples comprising reference images, instructions, executable Blender programs, proxy renderings, and target images. Through lightweight fine-tuning, pretrained generators learn to follow the proxy's composition while preserving the appearance of the corresponding reference subjects.

Experiments on DreamOmni2 and SconeEval demonstrate improvements in overall scores with both Qwen-Image-Edit-2511 and FLUX.2-klein-base-9B. On DreamOmni2, MAGPIE raises the FLUX backbone's overall score from 38.87\% to 67.08\%, and outperforms ordinary fine-tuning on the same reference--target pairs by 18.18 percentage points. The Qwen backbone likewise benefits beyond ordinary fine-tuning. These results support combining explicit scene construction with proxy-guided adaptation to improve reference-grounded image synthesis.

Our contributions are threefold:
\begin{itemize}
    \item We propose MAGPIE, a framework that instantiates reference-associated scene relationships in executable proxies for inspection and correction before appearance synthesis.
    \item We develop a structured harness for proxy construction and revision, and construct MAGPIE-Instruct-3K to teach generators to combine proxy structure with reference-specific appearance.
    \item We demonstrate improvements across two backbones and two benchmarks, including gains beyond ordinary generator fine-tuning.
\end{itemize}


\section{Related Work}
\label{sec:related_work}

\begin{figure*}[!ht]
  \centering
  \includegraphics[width=\linewidth]{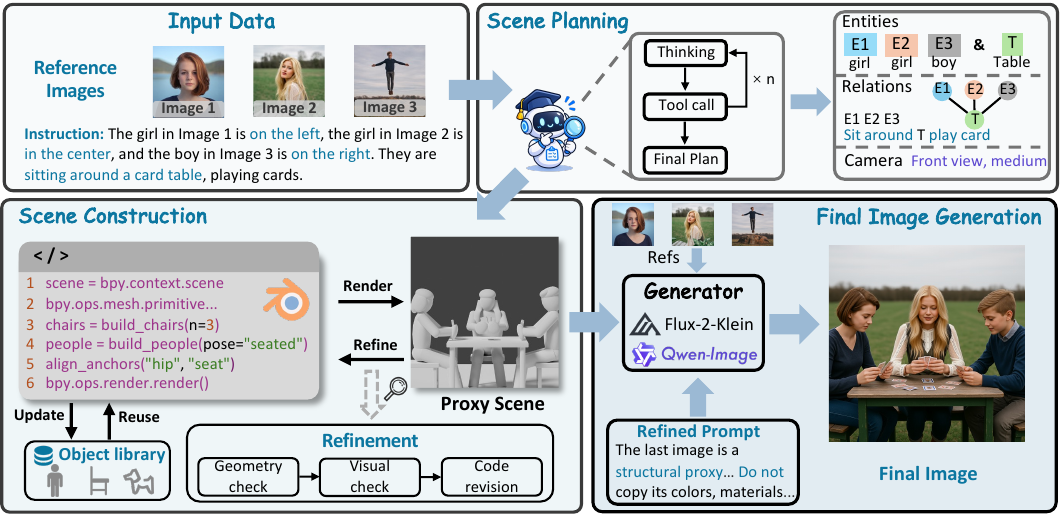}
  \caption{\textbf{Overview of MAGPIE.} The coding harness converts reference images and an instruction into an executable proxy scene through planning, procedural construction, and geometric and visual refinement. The generator combines the proxy's structure with reference-specific appearance. MAGPIE-Instruct-3K supports lightweight adaptation to this conditioning. }
  \label{fig:framework}
\end{figure*}

\subsection{Multi-Reference Image Generation}
Unlike conventional text-to-image generation~\cite{ICLR2024_081b0806,pmlr-v235-esser24a,NEURIPS2025_38fc6254,tong2026cof}, multi-reference image generation allows users to combine subjects and visual attributes drawn from different reference images following textual instructions~\cite{mou2025dreamo,wu2026omnigen2,wu2025qwen,deng2025bagel}. DreamOmni2~\cite{Xia_2026_CVPR} supports generation and editing with references to concrete subjects and abstract attributes, while Scone~\cite{Wang_2026_CVPR} distinguishes requested subjects from distractors. AutoRef~\cite{oshima2026autoref} further explores agentic generation by automatically optimizing the harness for reference processing, image synthesis, and output selection. A central challenge is preserving reference-specific appearance while arranging the selected subjects as instructed. This becomes more demanding when an instruction calls for multiple instances of the same subject. Each instance must retain the subject's appearance while being placed and posed according to the instruction.

\subsection{Structural Control for Image Generation}
Structured planning expresses compositional intent through semantic descriptions and spatial conditions. Textual reasoning in Re-Align~\cite{He_2026_CVPR} clarifies target-scene content and reference associations, while layout-based approaches such as LayoutGPT~\cite{feng2023layoutgpt} and LLM-grounded Diffusion~\cite{lian2023llm} specify object placement. In reference-conditioned generation, ContextGen~\cite{ICLR2026_9b01b794} and ConsistCompose~\cite{Shi_2026_CVPR} further associate subject identities with regions or coordinates. Richer conditions in ControlNet~\cite{zhang2023adding} and SeeThrough3D~\cite{agrawal2026seethrough3d} extend control to pose, depth, viewpoint, and occlusion. These approaches provide increasingly detailed structural guidance, but such guidance alone does not establish that all requested relations are jointly realized. Verifying the composition and localizing necessary revisions before final synthesis remains challenging.

\subsection{Executable Scene Representations}
Executable scene code allows a composition to be rendered, checked against its specification, and revised before final image synthesis. Existing approaches explore procedural scenes and programmatic sketches, as in SceneCraft~\cite{hu2024scenecraft}, CoCo~\cite{li2026coco} and GenClaw~\cite{ye2026genclaw}, or editable 3D scenes and layout refinement, as in MUSES~\cite{Ding_Zhuang_Li_Yue_Qiao_Wang_2025}, 3D Scratchpad~\cite{saha20263d}, and SpatialGuard~\cite{qian-etal-2026-spatialguard}. These methods primarily address text-driven scene construction, leaving the integration of user-provided subject identities into executable scenes less explored. To address this gap, MAGPIE grounds executable scene entities in the supplied references, enabling composition verification and revision while preserving reference-specific appearance in the final synthesis.

\section{Method}
\label{3_method}

Given reference images $\mathcal{I}=\{I_1,\ldots,I_N\}$ and an instruction $p$, MAGPIE generates an image that preserves reference-specific appearances while satisfying the requested composition. As illustrated in Fig.~\ref{fig:framework}, a multimodal coding harness first constructs and refines an executable Blender scene specifying entity selection, instance counts, spatial arrangement, and interactions. The image generator then uses the proxy rendering for structure and the original references for appearance, with explicit bindings between reference subjects and scene entities. We detail these stages below, followed by MAGPIE-Instruct-3K and lightweight adaptation for proxy-guided synthesis.

\subsection{Agentic Proxy Scene Construction}
\label{sec:scene_construction}

To instantiate the intended composition before image synthesis, we develop a multimodal coding harness that translates reference images and instructions into executable Blender scenes. The harness integrates reference-grounded scene planning, constrained procedural construction, and render-guided refinement, producing a structural proxy that captures the target entities and their spatial organization without requiring precise appearance reconstruction.

\paragraph{Reference-Grounded Scene Planning.}
The coding agent selects the requested subjects from the references and creates an instance-level scene specification $\mathcal{S}$ with entity counts, spatial and interaction relationships, and a camera configuration. Each entity has a unique identifier, reference associations, an approximate layout and pose, and a proxy category for procedural modeling. A relational graph encodes relative positioning, containment, support, and physical contact. We validate entity identifiers, reference associations, and relational endpoints before construction. Because the specification describes individual instances, it supports selective subject inclusion and multiple instances of the same reference subject with distinct compositional roles.

\paragraph{Constrained Procedural Scene Construction.}
Based on the scene specification, the coding agent generates a Blender program that instantiates the required entities and configures their geometry, placement, and camera viewpoint. To improve scene construction efficiency, we maintain a reusable library $\mathcal{A}$ of parameterized procedural modeling functions covering common entity categories, including articulated figures, animals, and structured objects. Each function exposes controllable parameters for geometry, scale, orientation, and pose. For entities not covered by the library, the agent constructs new procedural components using geometric primitives or custom meshes, which are subsequently incorporated into $\mathcal{A}$ for reuse in future scene construction. Guided by the assigned proxy categories, the agent generates an individual builder for each entity, together with a shared scene configuration. For entities covered by the library, each builder is required to invoke its designated procedural function. Before execution, we perform AST-based validation to ensure that all required builders are present and comply with their corresponding modeling interfaces.

\begin{figure*}[!ht]
  \centering
  \includegraphics[width=\linewidth]{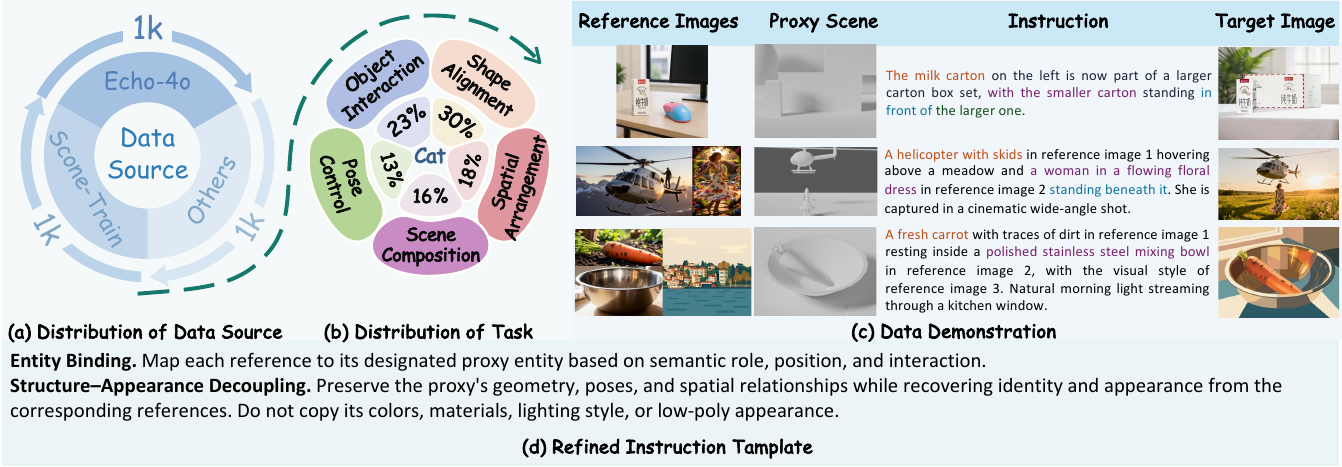}
  \caption{\textbf{Overview of MAGPIE-Instruct-3K.
} Across five compositional tasks, the dataset pairs reference images and instructions with proxy renderings and target images to teach generators to follow the proxy's structure while preserving reference-specific appearance.}
  \label{fig:data}
\end{figure*}

Independently constructed entities must also interact coherently. The components therefore use stable entity-specific mesh identifiers and shared world-space anchors to align interacting parts, such as a hand with an object's grasping region or a seated character with a supporting surface. Geometric feasibility checks, including articulated-limb reachability, identify inconsistent configurations during construction. Executing and rendering the program produces an initial proxy, while the mesh identifiers and anchors allow subsequent checks and revisions to target individual entities and interactions.

\paragraph{Geometry and Vision Guided Refinement.}
Although the initial program instantiates the planned scene, its rendered composition may still contain errors in entity visibility, spatial arrangement, or physical interactions. We therefore introduce a render guided refinement loop that combines geometric verification with multimodal visual review to identify discrepancies and guide code revision.

The geometric verifier examines the executed Blender meshes to assess whether the constructed scene satisfies the specified structural constraints. It first checks the presence and visibility of planned entities through mesh identifiers, projected bounding boxes, and ray-based visibility estimation. Camera-relative spatial relationships are then examined using projected positions and depth ordering to detect incorrect left-right or front-back arrangements. For interactions involving physical contact, the verifier evaluates surface distances between designated mesh components and checks the visibility of relevant contact regions. These measurements provide localized feedback on geometric inconsistencies in the rendered composition.

To detect semantic errors beyond these geometric measurements, a multimodal reviewer examines the proxy rendering and localized crops alongside the references, instruction, scene specification, and geometric feedback. It checks entity inventory, reference associations, relative placement, poses, and interactions. At round $k$, let $c^{(k)}$ denote the Blender program, $\mathcal{M}^{(k)}$ its executed meshes, and $D^{(k)}$ its rendering. The refinement loop is
\begin{equation}
\begin{aligned}
    g^{(k)} &= \operatorname{Geom}(\mathcal{S},\mathcal{M}^{(k)},D^{(k)}),\\
    f^{(k)} &= \operatorname{Review}(\mathcal{I},p,\mathcal{S},D^{(k)},g^{(k)}),\\
    c^{(k+1)} &= \operatorname{Revise}(c^{(k)},g^{(k)},f^{(k)}),\\
    D^{(k+1)} &= \operatorname{Render}(c^{(k+1)}),
\end{aligned}
\label{eq:proxy_refinement}
\end{equation}
where $g^{(k)}$ and $f^{(k)}$ are geometric and visual feedback, respectively. The agent uses both to revise geometry, placement, interaction anchors, and camera configuration; the resulting rendering is evaluated in the next round. The harness retains the best candidate according to the verification results and performs at most six refinement rounds, stopping early when all verification criteria are satisfied.

\paragraph{Cross-Example Harness Refinement.}
Instance-level revision corrects individual scenes; recurring errors also inform offline improvements to the harness. We construct proxies for over 200 randomly sampled training examples and compare their renderings with the corresponding target images. Common failure patterns guide changes to scene planning and code generation to improve initial proxy quality and reduce repeated repairs. Target images are used for this offline development; at inference, the harness receives only the references and instruction.

\subsection{Proxy-Guided Reference-Grounded Synthesis}
\label{sec:proxy_synthesis}

Building upon the proxy scene constructed in the preceding stage, we use its final rendering $D^\star$ to provide explicit structural guidance for reference-grounded image synthesis. The proxy specifies entity arrangement, approximate geometry, poses, and interactions, while the original references $\mathcal{I}$ supply identity and appearance details. We combine these inputs through
\begin{equation}
    \hat{Y}=G_{\theta+\Delta\theta}(\mathcal{I},D^\star,\tilde{p}),
    \label{eq:proxy_synthesis}
\end{equation}
where $\hat{Y}$ is the generated image, $G$ is the image generator, and $\theta$ and $\Delta\theta$ denote its pretrained parameters and LoRA updates, respectively. The proxy-aware prompt $\tilde{p}$ is derived from the instruction $p$ and the reference-to-entity bindings. It assigns each input its role: follow the proxy's placement, poses, and spatial relationships; preserve each subject's appearance from its bound reference; and avoid reproducing the proxy's simplified materials and colors.


\begin{table*}[!h]
  \caption{\textbf{Quantitative comparison on DreamOmni2.} Small green numbers show percentage-point changes in the three summary
  metrics relative to the corresponding backbone. Best scores in each group are highlighted in bold.}
  \label{tab:dreamomni2_gemini}
  \centering
  \begingroup
  \rmfamily
  \newcommand{\magpieup}[1]{\rlap{{\fontsize{7pt}{8pt}\selectfont\color{green!55!black}\,+#1}}}
  \newcommand{\magpiedown}[1]{\rlap{{\fontsize{7pt}{8pt}\selectfont\color{red!75!black}\,-#1}}}
  \newcommand{\magpiezero}[1]{\rlap{{\fontsize{7pt}{8pt}\selectfont\color{gray}\,\phantom{+}#1}}}
  \begin{tabular*}{\textwidth}{@{}l@{\extracolsep{\fill}}ccccccc@{\hspace{16pt}}}
    \toprule
    \multirow{2}{*}{\textbf{Method / Setting}}
    & \multicolumn{3}{c}{\textbf{Generation} $\uparrow$}
    & \multicolumn{3}{c}{\textbf{Editing} $\uparrow$}
    & \multirow{2}{*}{\textbf{Overall} $\uparrow$} \\
    \cmidrule(lr){2-4} \cmidrule(lr){5-7}
    & \textbf{Concrete} & \textbf{Abstract} & \textbf{All}
    & \textbf{Concrete} & \textbf{Abstract} & \textbf{All} & \\
    \midrule
    \rowcolor{BenchPeach}
    \multicolumn{8}{c}{\textbf{\textit{Proprietary Models}}} \\
    \midrule
    Gemini-2.5-flash-Image~\cite{google2025gemini} & 50.00 & 55.56 & 54.39 & 68.29 & 64.63 & 65.37 & 61.44 \\
    Seedream-4.0~\cite{seedream2025seedream} & \textbf{66.67} & \textbf{73.33} & \textbf{71.93} & \textbf{73.17} & \textbf{65.85} & \textbf{67.32} & \textbf{68.97} \\
    \midrule
    \rowcolor{BenchBlue}
    \multicolumn{8}{c}{\textbf{\textit{Open Weight Models}}} \\
    \midrule
    UNO~\cite{wu2025less} & 0.00 & 3.33 & 2.63 & 0.00 & 0.61 & 0.49 & 1.25 \\
    DreamO~\cite{mou2025dreamo} & 4.17 & 6.67 & 6.14 & 2.44 & 1.83 & 1.95 & 3.45 \\
    OmniGen2~\cite{wu2026omnigen2} & 20.83 & 10.00 & 12.28 & 21.95 & 4.27 & 7.80 & 9.40 \\
    FLUX.1 Kontext~\cite{labs2025flux1kontextflowmatching} & 25.00 & 5.56 & 9.65 & 4.88 & 1.83 & 2.44 & 5.02 \\
    Qwen-Image-Edit-2509~\cite{wu2025qwen} & 12.50 & 11.11 & 11.40 & 26.83 & 4.88 & 9.27 & 10.03 \\
    DreamOmni2~\cite{Xia_2026_CVPR} & 58.33 & 57.78 & 57.89 & 58.54 & 58.54 & 58.54 & 58.31 \\
    \midrule
    \rowcolor{BenchLavender}
    \multicolumn{8}{c}{\textbf{\textit{Different Base Models (MAGPIE)}}} \\
    \midrule
    \textbf{Qwen-Image-Edit-2511}~\cite{wu2025qwen} & 62.50 & 41.11 & 45.61 & 56.10 & 35.98 & 40.00 & 42.01 \\
    \quad + MAGPIE & 62.50 & \textbf{68.89} & 67.54\magpieup{21.93} & 63.41 & 62.20 & 62.44\magpieup{22.44} & 64.26\magpieup{22.25} \\
    \textbf{FLUX.2-klein-base-9B}~\cite{flux-2-2025} & 70.83 & 26.67 & 35.96 & 68.29 & 33.54 & 40.49 & 38.87 \\
    \quad + MAGPIE & \textbf{70.83} & 66.67 & \textbf{67.54}\magpieup{31.58} & \textbf{73.17} & \textbf{65.24} & \textbf{66.83}\magpieup{26.34} & \textbf{67.08}\magpieup{28.21} \\
    \bottomrule
  \end{tabular*}
  \endgroup
\end{table*}

\begin{table*}[t]
  \caption{\textbf{Quantitative comparison on SconeEval.}
  Scores range from 0 to 10 (higher is better). COM and DIS denote composition
  and distinction. Best scores in each group are highlighted in bold.}
  \label{tab:sconeeval_comparison}
  \centering
  \begingroup
  \rmfamily
  \setlength{\tabcolsep}{1mm}
  \resizebox{\textwidth}{!}{%
    \begin{tabular}{@{}l*{13}{c}@{}}
      \toprule
      \multirow{3}{*}{\textbf{Method / Setting}}
      & \multicolumn{2}{c}{\textbf{Composition} $\uparrow$}
      & \multicolumn{4}{c}{\textbf{Distinction} $\uparrow$}
      & \multicolumn{4}{c}{\textbf{Distinction \& Composition} $\uparrow$}
      & \multicolumn{3}{c}{\textbf{Average} $\uparrow$} \\
      \cmidrule(lr){2-3}
      \cmidrule(lr){4-7}
      \cmidrule(lr){8-11}
      \cmidrule(l){12-14}
      & \textbf{Single} & \textbf{Multi}
      & \multicolumn{2}{c}{\textbf{Cross}}
      & \multicolumn{2}{c}{\textbf{Intra}}
      & \multicolumn{2}{c}{\textbf{Cross}}
      & \multicolumn{2}{c}{\textbf{Intra}}
      & & & \\
      \cmidrule(lr){2-2}
      \cmidrule(lr){3-3}
      \cmidrule(lr){4-5}
      \cmidrule(lr){6-7}
      \cmidrule(lr){8-9}
      \cmidrule(lr){10-11}
      & \textbf{COM} & \textbf{COM}
      & \textbf{COM} & \textbf{DIS}
      & \textbf{COM} & \textbf{DIS}
      & \textbf{COM} & \textbf{DIS}
      & \textbf{COM} & \textbf{DIS}
      & \textbf{COM} & \textbf{DIS} & \textbf{Overall} \\
      \midrule
      \rowcolor{BenchPeach}
      \multicolumn{14}{c}{\textbf{\textit{Proprietary Models}}} \\
      \midrule
      Gemini-2.5-Flash-Image~\cite{google2025gemini}
      & 8.87 & 7.94 & 9.12 & \textbf{9.15} & 9.00 & 8.50
      & 8.27 & \textbf{8.87} & 8.17 & 8.85 & 8.56 & 8.84 & 8.70 \\
      GPT-4o~\cite{openai2025gpt4o}
      & \textbf{8.92} & \textbf{8.51} & \textbf{9.18} & 8.55
      & \textbf{9.45} & \textbf{9.01} & \textbf{8.83} & 8.49
      & \textbf{8.99} & \textbf{9.56} & \textbf{8.98}
      & \textbf{8.90} & \textbf{8.94} \\
      \midrule
      \rowcolor{BenchBlue}
      \multicolumn{14}{c}{\textbf{\textit{Open-Weight Models}}} \\
      \midrule
      UNO~\cite{wu2025less}
      & 7.53 & 5.38 & 7.27 & 7.90 & 6.76 & 6.53
      & 5.27 & 7.02 & 5.61 & 6.27 & 6.31 & 6.93 & 6.62 \\
      Qwen-Image-Edit-2509~\cite{wu2025qwen}
      & 8.54 & 6.85 & 8.85 & 8.57 & 7.32 & 6.86
      & 7.53 & 8.13 & 7.49 & 7.02 & 7.76 & 7.65 & 7.70 \\
      BAGEL~\cite{deng2025bagel}
      & 7.14 & 5.55 & 7.49 & 7.95 & 6.93 & 6.21
      & 6.44 & 7.38 & 6.87 & 7.27 & 6.74 & 7.20 & 6.97 \\
      OmniGen2~\cite{wu2026omnigen2}
      & 8.00 & 6.59 & 8.31 & 8.99 & 6.99 & 6.80
      & 7.28 & 8.30 & 7.14 & 7.13 & 7.39 & 7.81 & 7.60 \\
      Echo-4o~\cite{ye2025echo}
      & 8.58 & 7.73 & 8.36 & 8.33 & 7.74 & 7.18
      & 7.87 & 8.72 & 8.01 & 8.33 & 8.05 & 8.14 & 8.09 \\
      Scone~\cite{Wang_2026_CVPR}
      & 8.52 & 7.40 & 8.98 & 9.73 & 7.97 & 7.74
      & 8.20 & 9.25 & 8.21 & 8.44 & 8.21 & 8.79 & 8.50 \\
      \midrule
      \rowcolor{BenchLavender}
      \multicolumn{14}{c}{\textbf{\textit{Different Base Models (MAGPIE)}}} \\
      \midrule
      \textbf{Qwen-Image-Edit-2511}~\cite{wu2025qwen}
      & 8.73 & 7.65 & 8.81 & 8.36 & 8.22 & 7.71
      & 7.99 & 8.89 & 7.88 & 7.95 & 8.21 & 8.23 & 8.22 \\
      \quad + MAGPIE
      & 8.44 & \textbf{8.11} & 8.83 & 9.56 & 8.71 & 8.85
      & \textbf{8.69} & \textbf{9.57} & \textbf{8.83} & 9.08
      & 8.60 & 9.26 & 8.93 \\
      \textbf{FLUX.2-klein-base-9B}~\cite{flux-2-2025}
      & \textbf{8.69} & 8.03 & \textbf{9.11} & 8.97 & 7.39 & 6.75
      & 8.22 & 8.59 & 7.75 & 7.42 & 8.20 & 7.93 & 8.07 \\
      \quad + MAGPIE
      & 8.67 & 7.93 & 9.07 & \textbf{9.74}
      & \textbf{9.17} & \textbf{9.33} & 8.45 & 9.29 & 8.76
      & \textbf{9.35} & \textbf{8.68} & \textbf{9.43} & \textbf{9.05} \\
      \bottomrule
    \end{tabular}%
  }
  \endgroup
\end{table*}

\subsection{MAGPIE-Instruct-3K and Model Adaptation}
\label{sec:model_adaptation}

Although proxy renderings provide explicit structural guidance, pretrained image generators may not interpret them as compositional constraints. As shown in Tab.~\ref{tab:magpie_ablation}, introducing proxy guidance into Qwen-Image-Edit-2511 without adaptation substantially degrades generation quality, motivating adaptation to this form of structural conditioning. However, existing multi-reference generation datasets generally lack paired executable proxy scenes and target images.

We therefore construct MAGPIE-Instruct-3K using the scene construction harness. Source examples come primarily from Echo-4o~\cite{ye2025echo} and the Scone~\cite{Wang_2026_CVPR} training set, supplemented by Pico-Banana~\cite{qian2026pico}, MICo-150K~\cite{wei2026mico}, and other public multi-reference generation datasets. Source details are provided in the supplementary material. We retain examples with identifiable entities and requirements expressible through proxy geometry, such as shape, poses and spatial arrangements. During construction, the coding agent receives the target image to align the proxy with the intended composition. We then use Qwen3.8-27B~\cite{qwen38} as a multimodal judge to filter proxy–target pairs based on entity inclusion and counts, spatial arrangements, poses and interactions, and framing and visibility. The evaluation prompt appears in the supplementary Material. The dataset contains 3187 examples, each comprising references, an instruction, an executable Blender program, a proxy rendering, and a target image. Fig.~\ref{fig:data} presents the source and task distributions, representative examples, and the proxy-aware prompt template.

Using this dataset, we adapt pretrained image generators through lightweight low-rank adaptation (LoRA), enabling them to leverage proxy-based structural guidance while preserving reference-specific appearances. Following the flow-matching formulation, let $z_0=\mathcal{E}(Y)$ denote the latent representation of the target image $Y$, and let $\epsilon\sim\mathcal{N}(0,I)$ denote Gaussian noise. The training objective is
\begin{equation}
\begin{aligned}
    z_t &= (1-t)z_0+t\epsilon,\\
    \mathcal{L}_{\mathrm{FM}}
    &=\mathbb{E}_{t,\epsilon,\mathcal{D}}
    \left[
    \left\|
    v_{\theta+\Delta\theta}(z_t,t;\mathcal{I},D^\star,\tilde{p})
    -(\epsilon-z_0)
    \right\|_2^2
    \right],
\end{aligned}
\label{eq:proxy_flow_matching}
\end{equation}
where $\mathcal{D}$ denotes MAGPIE-Instruct-3K, $D^\star$ is the proxy rendering, and $\tilde{p}$ is the proxy-aware synthesis prompt introduced in the preceding section. Here, $\theta$ represents the frozen pretrained parameters, while $\Delta\theta$ denotes the trainable LoRA updates.

\begin{figure*}[!ht]
  \centering
  \includegraphics[width=\linewidth]{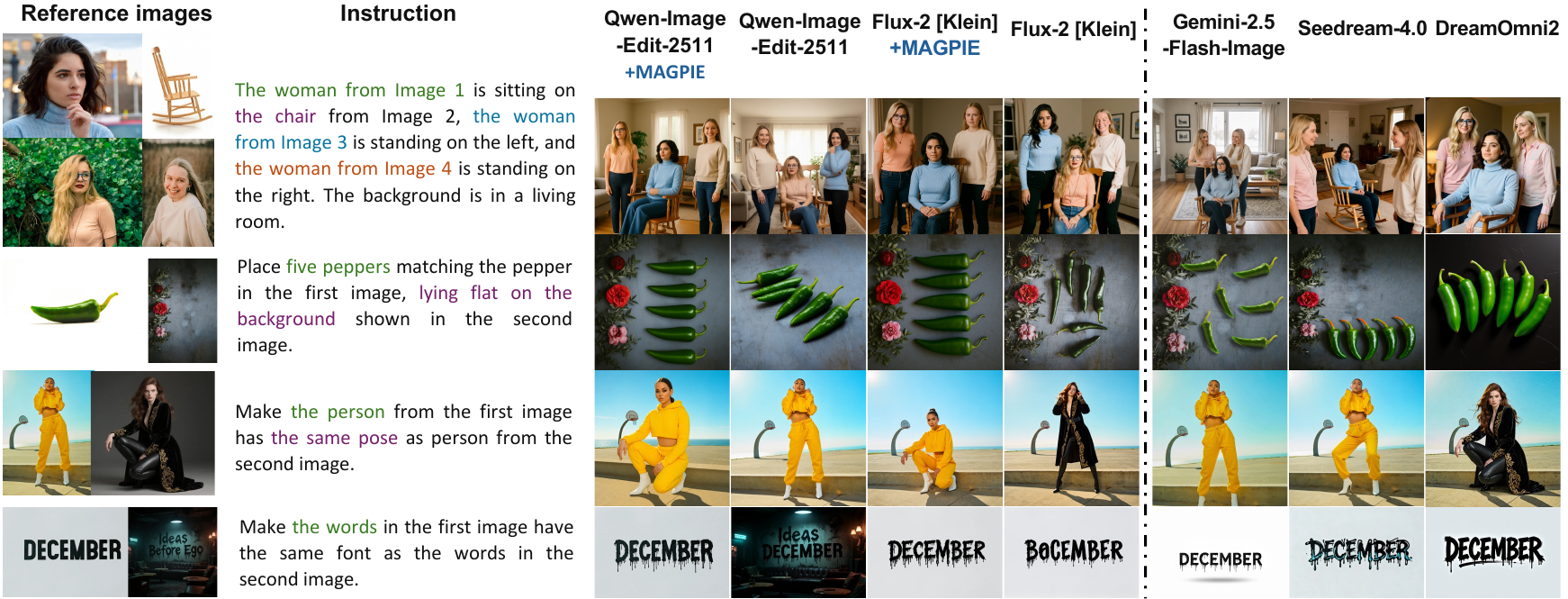}
  \caption{Qualitative comparison of existing models on DreamOmni2 benchmark.}
  \label{fig:case_study}
\end{figure*}
\section{Experiments}
\label{4_experiment}

We evaluate MAGPIE with two image generation backbones on multi-reference generation and editing benchmarks, then examine the contributions of generator adaptation, scene representation, and proxy refinement.

\subsection{Experimental Setting}
\label{sec:experimental_setting}

\paragraph{Training.}
We adapted Qwen-Image-Edit-2511~\cite{wu2025qwen} and FLUX.2 [klein]~\cite{flux-2-2025}, referred to as Qwen and FLUX, respectively, on the 3187 examples in MAGPIE-Instruct-3K. We used Qwen3.8-flash~\cite{qwen3.8flashnext} for agentic scene construction and visual review during both offline data construction and inference. We trained rank-32 LoRA adapters for five epochs with a learning rate of $10^{-4}$ and a batch size of one, using images up to $1024\times1024$ resolution. The pretrained generator parameters remained frozen. At inference, the agent receives only the references and instruction, and the harness performs at most six refinement rounds with early stopping, as described in Sec.~\ref{sec:scene_construction}.

\paragraph{Evaluation.}
We evaluate on DreamOmni2~\cite{Xia_2026_CVPR} and SconeEval~\cite{Wang_2026_CVPR} using their official evaluation scripts, scoring prompts, and aggregation procedures. DreamOmni2 covers generation and editing with concrete objects and abstract attributes. Its evaluation measures whether the output successfully satisfies the multimodal instruction; we report success rates for both categories, their aggregate within each task, and an overall score, all as percentages. SconeEval evaluates composition, distinction, and their combination, including cross-category and intra-category subject selection. Its composition score (COM) combines prompt following and subject consistency through their per-example geometric mean. Its distinction score (DIS) averages subject-selection accuracy and F1, rescaled to the same 0--10 range. We report the benchmark's subtask scores and aggregate COM, DIS, and Overall, with higher values indicating better performance. 

\subsection{Quantitative and Qualitative Evaluation}
\label{sec:main_results}

\paragraph{Quantitative evaluation.}
On DreamOmni2, MAGPIE improves both generation and editing success with either backbone (Tab.~\ref{tab:dreamomni2_gemini}). Qwen gains 22.25 percentage points in overall success over the original backbone. The gain is larger for FLUX, whose success rate rises from 38.87\% to 67.08\%, a 28.21-point increase. Both variants outperform the strongest open-weight baseline, DreamOmni2~\cite{Xia_2026_CVPR}, and exceed the 61.44\% overall success rate of the proprietary Gemini-2.5-Flash-Image.

On SconeEval, MAGPIE improves both composition and subject distinction with either backbone (Tab.~\ref{tab:sconeeval_comparison}). With FLUX, the average composition score rises from 8.20 to 8.68 and the average distinction score from 7.93 to 9.43. These gains bring the overall score to 9.05, the highest among all models compared, ahead of the proprietary GPT-4o (8.94) and Gemini 2.5 Flash Image (8.70). Together, these results support using executable 3D scene proxies to guide composition in multi-reference image generation.

\paragraph{Qualitative evaluation.}
Figure~\ref{fig:case_study} compares Qwen-Image-Edit-2511 and FLUX.2 [klein], each with and without MAGPIE, on DreamOmni2Bench. We also include results from Gemini 2.5 Flash Image, Seedream 4.0, and DreamOmni2. With MAGPIE, both backbones follow the requested compositions and edits more closely while preserving the visual identity of the referenced subjects. This consistency is visible even when the instructions call for a change in a subject's position or pose.

\begin{table}[!t]
  \caption{\textbf{Ablation study of MAGPIE and SFT on DreamOmni2Bench.} We report aggregated generation, editing, and overall scores. Bold indicates the best result within each backbone group.}
  \label{tab:magpie_ablation}
  \centering
  \small
  \setlength{\tabcolsep}{3pt}
  \renewcommand{\arraystretch}{1.10}
  \begin{tabular*}{\columnwidth}{@{}l@{\extracolsep{\fill}}ccc@{}}
    \toprule
    \textbf{Method / Setting} & \textbf{Gen. All} $\uparrow$ & \textbf{Edit. All} $\uparrow$ & \textbf{Overall} $\uparrow$ \\
    \midrule
    \textbf{Qwen-Image-Edit-2511} & 45.61 & 40.00 & 42.01 \\
    \quad + Proxy only & 35.09 & 26.34 & 29.47 \\
    \quad + Standard SFT & 42.11 & 46.34 & 44.83 \\
    \quad + MAGPIE & \textbf{67.54} & \textbf{62.44} & \textbf{64.26} \\
    \midrule
    \textbf{FLUX.2-klein-base-9B} & 35.96 & 40.49 & 38.87 \\
    \quad + Proxy only & 40.35 & 49.27 & 46.08 \\
    \quad + Standard SFT & 46.49 & 50.24 & 48.90 \\
    \quad + MAGPIE & \textbf{67.54} & \textbf{66.83} & \textbf{67.08} \\
    \bottomrule
  \end{tabular*}
\end{table}

\begin{table}[!t]
  \caption{\textbf{Comparison of different scene planning strategies on DreamOmni2Bench.} We report generation scores on concrete and abstract tasks and their sample-weighted aggregate. Bold numeric values denote MAGPIE.}
  \label{tab:planning_ablation}
  \centering
  \small
  \setlength{\tabcolsep}{3pt}
  \renewcommand{\arraystretch}{1.10}
  \begin{tabular*}{\columnwidth}{@{}l@{\extracolsep{\fill}}ccc@{}}
    \toprule
    \textbf{Method / Setting} & \textbf{Gen. Con} $\uparrow$ & \textbf{Gen. Abs} $\uparrow$ & \textbf{Gen. All} $\uparrow$ \\
    \midrule
    \textbf{Qwen-Image-Edit-2511} & 62.50 & 41.11 & 45.61 \\
    \quad + Text Scene & 54.17 & 54.44 & 54.39 \\
    \quad + 2D Scene & 50.00 & 56.67 & 55.26 \\
    \quad + 3D Scene (MAGPIE) & \textbf{62.50} & \textbf{68.89} & \textbf{67.54} \\
    \midrule
    \textbf{FLUX.2-klein-base-9B} & 70.83 & 26.67 & 35.96 \\
    \quad + Text Scene & 58.33 & 58.89 & 58.77 \\
    \quad + 2D Scene & 33.33 & 51.11 & 47.37 \\
    \quad + 3D Scene (MAGPIE) & \textbf{70.83} & \textbf{66.67} & \textbf{67.54} \\
    \bottomrule
  \end{tabular*}
\end{table}

\subsection{Ablation Study}
\label{sec:ablation}

\paragraph{Effect of proxy-conditioned adaptation.}
We disentangle the effects of additional training data, inference-time proxy guidance, and proxy-conditioned adaptation in Tab.~\ref{tab:magpie_ablation}. Standard SFT uses the same references,  
instructions, and target images as MAGPIE but excludes proxy renderings during both training and inference. It improves the overall score from 42.01\% to 44.83\% for Qwen and from 38.87\% to 48.90\% for FLUX, confirming that the training pairs themselves are beneficial. Conditioning on proxies during both adaptation and inference further raises the scores to 64.26\% and 67.08\%, outperforming standard SFT by 19.43 and 18.18 percentage points, respectively. In contrast, supplying the proxy only at inference without adaptation produces inconsistent effects: it substantially degrades Qwen to 29.47\%, while improving FLUX to 46.08\%. These results show that proxy renderings are not universally effective as zero-shot guidance; their structural benefit is realized most reliably when the generator is trained to interpret them.

\paragraph{Effect of scene representation.}We compare three representations of the scene plan (Tab.~\ref{tab:planning_ablation}). Text Scene uses the entity-level plan produced by the first stage of the MAGPIE harness as textual guidance, while 2D Scene converts that plan into labeled bounding boxes. For each variant, we used this pipeline to construct training examples from MAGPIE-Instruct-3K and trained a separate LoRA adapter for each backbone. At inference, the same pipeline produces the corresponding guidance from the input references and instruction. With a rendered Blender proxy as guidance, 3D Scene achieves the highest aggregate generation score on both backbones (67.54\%), exceeding the best alternative planning variant by 12.28 percentage points on Qwen and 8.77 on FLUX.

\begin{figure}[!t]
  \centering
  \includegraphics[width=\linewidth]{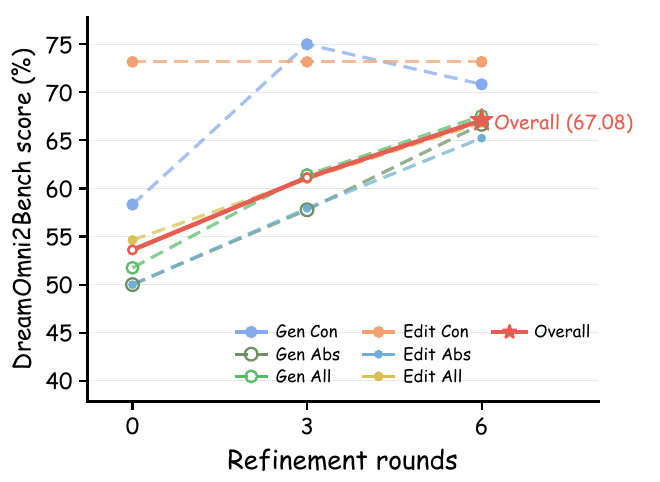}
  \caption{\textbf{Generation and editing performance across refinement rounds.} We visualize the progressive improvement in MAGPIE's aggregate generation and editing scores on DreamOmni2Bench across 0, 3, and 6 rounds of proxy refinement.}
  \label{fig:ablation_proxy_quality}
\end{figure}

\paragraph{Effect of proxy refinement.}
Using the adapted FLUX backbone, we evaluate generation and editing performance by setting the maximum number of proxy refinement rounds to 0, 3, or 6. At zero rounds, the generator uses the initial proxy before render-guided refinement. Figure~\ref{fig:ablation_proxy_quality} reports DreamOmni2Bench scores for concrete and abstract examples in both generation and editing, along with the aggregate score for each task and the overall score. Both task aggregates rise across the three budgets, and the overall score increases from 53.61\% at zero rounds to 67.08\% at six rounds. The supplementary material also provides proxy construction reliability statistics, qualitative comparisons across refinement rounds, proxy quality assessments, and runtime analysis.

\section{Conclusion}
\label{5_conclusion}
In this work, we introduce MAGPIE, a framework for multi-reference image generation that represents the requested composition as an executable 3D scene. A coding agent arranges the selected subjects in this scene, checks its geometry and rendered view against the instruction, and revises the scene as needed. We also curate MAGPIE-Instruct-3K for lightweight fine-tuning so the generator learns to follow the refined rendering for composition and the original references for appearance. Experiments on DreamOmni2 and SconeEval show that MAGPIE improves overall performance with both Qwen and FLUX. We hope this work encourages further study of how executable scene representations can give users finer control over multi-reference image generation.

\section{LLM Usage Statement}
\label{sec:llm usage statement}
Qwen3.8-Flash-Next powered the scene construction and refinement agent, while Qwen3.8-27B supported quality filtering during the construction of MAGPIE-Instruct-3K. Gemini-2.5-Flash and GPT-4.1 served as judges for evaluation on DreamOmni2Bench and SconeEval, respectively. AI tools also assisted with implementation and experimental workflows, as well as language polishing and clarity improvements during manuscript preparation. The authors made all research decisions and drew all conclusions, and take full responsibility for the methodology, experimental design, validation and interpretation of results, and final content of the manuscript.
{
    \small
    \bibliographystyle{ieeenat_fullname}
    \bibliography{main}
}

\clearpage
\setcounter{page}{1}
\maketitlesupplementary

\definecolor{PromptPurple}{RGB}{128,0,128}
\definecolor{PromptPurpleDark}{RGB}{102,0,102}
\newtcolorbox{promptstylebox}[1]{
  enhanced,
  colback=white,
  colframe=PromptPurple,
  boxrule=1.0pt,
  arc=2.5mm,
  left=4mm,
  right=4mm,
  top=4mm,
  bottom=3mm,
  fontupper=\small\itshape,
  title=#1,
  colbacktitle=PromptPurpleDark,
  coltitle=white,
  fonttitle=\bfseries,
  boxed title style={
    arc=1.5mm,
    boxrule=0pt,
    left=4mm,
    right=4mm,
    top=1.2mm,
    bottom=1.2mm
  },
  attach boxed title to top left={xshift=9mm,yshift*=-\tcboxedtitleheight/2},
}

\section{Training Dataset Details}
\label{sec:rationale}

MAGPIE-Instruct-3K contains 3,187 examples for proxy-conditioned
image generator adaptation, drawn from subject-driven generation,
multi-reference composition, and reference-guided editing datasets.
Construction comprises source selection, executable proxy construction,
and structural filtering of proxy--target pairs. We describe these
stages and the resulting task categories below. The corresponding
prompt templates accompany the descriptions of each assessment stage.

\subsection{Data Sources}
\label{sec:data_sources}

\paragraph{Echo-4o-Image.}
Echo-4o-Image~\cite{ye2025echo} is a GPT-4o-generated dataset
covering multi-reference generation, surreal fantasy generation,
and complex instruction following. Its multi-reference subset
pairs visual references and textual instructions with generated
target images, supporting diverse reference-conditioned
compositions.

\paragraph{Scone-Train.}
Scone-Train~\cite{Wang_2026_CVPR} supports subject-driven
generation in both single-candidate and multi-candidate
settings. In the multi-candidate setting, each reference image
may contain multiple candidate subjects, requiring the model
to identify those specified by the instruction and compose
them while preserving their identities.

\paragraph{Pico-Banana via RCEdit.}
Pico-Banana-400K~\cite{qian2026pico} provides image-editing
triplets consisting of a source image, an instruction, and
an edited target. We use its reference-conditioned adaptation
provided by RCEdit~\cite{zhang2026rcedit}, which augments the original
editing pairs with reference images and corresponding
instructions. These examples specify editing requirements
through both textual and visual conditions.

\paragraph{MICo-150K.}
MICo-150K~\cite{wei2026mico} supports multi-image composition
across human-centric, object-centric, and human--object
interaction tasks. It also includes a
Decomposition-and-Recomposition subset, in which complex
images are decomposed into component references for
subsequent recomposition.

\paragraph{OmniRef-training.}
OmniRef-training~\cite{huang2026scaling} supports generation
conditioned on heterogeneous references that specify subjects,
backgrounds, poses, lighting, and styles. Different reference
images can therefore provide distinct types of guidance
within the same example.

\paragraph{MacroData.}
MacroData~\cite{chen2026macro}, introduced in MACRO, covers
customization, illustration, spatial reasoning, and temporal
prediction. Its customization component focuses on
reference-based composition, while its spatial component
addresses novel-view generation.

\paragraph{UNO-1M.}
UNO-1M~\cite{wu2025less} provides image pairs constructed
through an in-context generation pipeline. It emphasizes
subject consistency across different contexts and includes
captions and visual consistency annotations for
subject-driven generation.

\subsection{Data Selection for Proxy Construction}
\label{sec:source_selection}

We sample candidate examples from the source datasets and assess
their suitability for proxy construction using Qwen3.8-27B,
conditioned on the references, instruction, and target image.
Selection focuses on whether the requested composition can be
represented by identifiable entities and simplified geometry,
including object shapes, instance counts, spatial arrangements,
poses, and interactions. Accepted examples must provide clear
reference-to-subject correspondence and sufficient visible evidence
to assess the relevant structural requirements.

Appearance-only transformations, such as texture or style changes
without a meaningful geometric requirement, are excluded. Mixed
requirements remain eligible when their structural component can
be expressed by a proxy. We also check the target for evident
conflicts with the instruction or referenced subjects, ensuring
that it provides a usable composition target for construction.
Figure~\ref{fig:prompt_source_selection} gives the selection prompt.

\begin{figure*}[!t]
\centering
\begin{minipage}{\textwidth}
\begin{promptstylebox}{Prompt 1: Source Example Selection}

\textbf{Role.}
You assess whether an image-generation example is suitable
for constructing a simplified 3D proxy scene.

\textbf{Input.}
Reference images, a generation or editing instruction,
and a target image. The target is a construction
reference and must itself be checked for consistency.

\textbf{Task.}
Determine whether the example contains a clear structural
composition that can be represented by coarse geometry
and evaluated in a rendered view.

\textbf{Selection Criteria.}
\begin{enumerate}[leftmargin=*,nosep]
  \item \textbf{Identifiable subjects.}
  The subjects required by the instruction can be identified
  and associated with the relevant references. Additional
  candidate subjects in a reference are allowed when the
  requested selection is unambiguous.

  \item \textbf{Geometric relevance.}
  The example contains a meaningful requirement involving
  shape, instance count, spatial arrangement, pose, support,
  containment, or contact. A simplified proxy should be
  able to express this requirement without reproducing
  detailed textures or materials.

  \item \textbf{Valid target.}
  The target depicts the requested subjects and composition
  without clear contradictions in identity, count, or the
  required relationships.

  \item \textbf{Observable structure.}
  The target provides sufficient evidence to assess the
  relevant arrangement, pose, or interaction. Reject cases
  where essential evidence is unavailable because of
  ambiguity or severe occlusion.
\end{enumerate}

\textbf{Decision Rules.}
Accept only when all four criteria are satisfied.
Reject appearance-only transformations without a meaningful
structural requirement. Do not reject a sample merely because
it also includes appearance conditions or requires complex
geometry. Judge representability, not ease of construction.
Do not invent hidden objects, contacts, or poses; mark a
criterion as false when the available evidence is insufficient.

\textbf{Output Format.}
Return raw JSON only. Each criterion is a Boolean;
\texttt{accept} is true only when all four are true.
Provide a short evidence-based reason.

{\ttfamily\footnotesize
\begin{verbatim}
{
  "identifiable_subjects": true,
  "geometric_relevance": true,
  "valid_target": true,
  "observable_structure": true,
  "accept": true,
  "reason": "Brief evidence supporting the decision."
}
\end{verbatim}
}

\end{promptstylebox}
\end{minipage}
\caption{Prompt template for selecting source examples
suitable for executable proxy construction. The assessment
uses the references, instruction, and target image before
a proxy is constructed.}
\label{fig:prompt_source_selection}
\end{figure*}

\subsection{Proxy Construction and Pair Filtering}
\label{sec:proxy_pair_filtering}

For each selected example, the scene construction harness produces
an executable Blender program and a proxy rendering. During this
stage, the coding agent receives the target image alongside
the references and instruction to align the proxy with the training
composition. Target images are used only during dataset construction;
at inference, the harness receives only the references and instruction.

We then assess proxy--target consistency using Qwen3.8-27B along
four dimensions: entity inclusion and counts, spatial arrangements,
poses and interactions, and framing and visibility. The assessment
checks whether the proxy provides structural guidance consistent
with the target, while allowing coarse shapes and simplified
articulation. Differences in texture, color, material, and other
appearance details are excluded. Figure~\ref{fig:prompt_pair_filtering}
provides the filtering prompt. Each retained example comprises
reference images, an instruction, an executable Blender program,
a proxy rendering, and a target image.

\begin{figure*}[!t]
\centering
\begin{minipage}{\textwidth}
\begin{promptstylebox}{Prompt 2: Proxy--Target Pair Filtering}

\textbf{Role.}
You assess whether a simplified proxy rendering provides
reliable structural guidance for its paired target image.

\textbf{Input.}
Reference images, a generation or editing instruction,
a target image, and a proxy rendering. The references
identify the requested subjects, the instruction specifies
the requirements, and the target depicts the composition
to be matched.

\textbf{Task.}
Evaluate structural agreement between the proxy and target
with respect to the requested composition.

\textbf{Evaluation Criteria.}
\begin{enumerate}[leftmargin=*,nosep]
  \item \textbf{Entity inclusion and counts.}
  The proxy contains the required subject instances with
  the correct counts. Reject missing, duplicated, or
  additional task-relevant instances.

  \item \textbf{Spatial arrangements.}
  Relative positions, depth ordering, orientation, and
  scale agree with the target where they affect the
  requested composition. Check left/right, above/below,
  front/behind, and containment when applicable.

  \item \textbf{Poses and interactions.}
  Task-relevant body configurations, object orientations,
  and interactions agree with the target. Check required
  grasping, support, contact, and containment relations;
  mere proximity is insufficient when contact is required.

  \item \textbf{Framing and visibility.}
  The proxy preserves the compositionally relevant framing
  and subject visibility. Reject occlusion, or
  viewpoint differences that hide required instances or
  substantially change the intended composition.
\end{enumerate}

\textbf{Evaluation Boundaries.}
Allow coarse geometry and simplified articulation when the
required structure remains clear. Ignore differences in
color, texture, material, lighting, facial detail, and
rendering realism. Do not require pixel-level alignment or
exact background reconstruction. Assess only visible
evidence; do not assume that a hidden relation is correct.

\textbf{Decision Rules.}
Assign each criterion \texttt{pass}, \texttt{fail}, or
\texttt{uncertain}. A criterion passes when its applicable
requirements are satisfied; it also passes if no such
requirement applies. Use \texttt{uncertain} when a required
property cannot be assessed from the inputs.
Accept only when all four criteria pass.

\textbf{Output Format.}
Return raw JSON only, with a short explanation identifying
the decisive agreement, mismatch, or missing evidence.

{\ttfamily\footnotesize
\begin{verbatim}
{
  "entities_and_counts": "pass",
  "spatial_arrangement": "pass",
  "pose_and_interaction": "pass",
  "framing_and_visibility": "pass",
  "accept": true,
  "reason": "Brief evidence supporting the decision."
}
\end{verbatim}
}

\end{promptstylebox}
\end{minipage}
\caption{Prompt template for filtering proxy--target pairs
by structural consistency. Appearance differences caused
by the simplified proxy representation are excluded
from the assessment.}
\label{fig:prompt_pair_filtering}
\end{figure*}

\subsection{Data Categorization}
\label{sec:data_categorization}

We organize the retained examples into five categories according
to their principal compositional requirement:
\textit{Shape Alignment}, concerning coarse object form and proportions;
\textit{Object Interaction}, concerning actions and functional
relationships between entities, including human--human and
human--object interactions;
\textit{Spatial Arrangement}, concerning relative positions,
depth ordering, and spatial containment;
\textit{Scene Composition}, concerning subject selection,
instance counts, and the assembly of subjects and backgrounds;
and \textit{Pose Control}, concerning a subject's posture and
articulated configuration.

Although instructions may combine several requirements, each
example receives one primary category based on its dominant
requested operation. Reference and target images are used to
resolve visual referents. Object Interaction concerns how entities
act together, whereas Pose Control concerns an individual subject's
configuration and Spatial Arrangement concerns relative placement.
Figure~\ref{fig:prompt_data_categorization} presents the categorization
prompt. We manually inspect the 338 examples flagged as ambiguous
and assign their final labels using the same category definitions.

\begin{table*}[!t]
  \caption{\textbf{Operational constraints for proxy verification.}
  Tests apply to the entities and relations declared in the scene
  specification. Image coordinates are normalized to $[0,1]$.
  Entity visibility measures occlusion by other entities, whereas
  exact-part visibility distinguishes individual meshes.
  Visual review complements these approximate geometric diagnostics.}
  \label{tab:operational_verification}
  \centering
  \begingroup
  \rmfamily
  \small
  \setlength{\tabcolsep}{3pt}
  \renewcommand{\arraystretch}{1.10}
  \begin{tabular*}{\textwidth}{@{}>{\raggedright\arraybackslash}p{0.20\textwidth}@{\extracolsep{\fill}}>{\raggedright\arraybackslash}p{0.76\textwidth}@{}}
    \toprule
    \textbf{Constraint} & \textbf{Operational rule} \\
    \midrule
    Inventory and ownership &
    Each planned entity has at least one executed mesh with its
    \texttt{entity\_id\_} prefix. Declared contact endpoints must map
    to meshes of the specified subject and target. \\
    Surface contact &
    For \texttt{touch}, \texttt{hold}, \texttt{support}, and
    \texttt{wear}, sample vertex-to-surface distances in both
    directions using bounding volume hierarchies (BVHs). The minimum
    sampled distance must satisfy $\widehat d\leq\tau$, with
    $\tau=\max\{10^{-5},\,0.025L,\,\min(2p,0.15L)\}$.
    Here $L$ is the smaller world-space bounding-box diagonal
    and $p$ is the camera-derived world-space size of one render pixel
    at the contact. \\
    Framing &
    For each frame-required entity, the union of projected mesh
    bounds must lie within $[-0.015,1.015]^2$, and no mesh vertex
    may lie behind the camera. \\
    Entity visibility &
    Cast a $20\!\times\!20$ ray grid over the clipped projected
    bounding box. Among rays hitting the entity in isolation,
    the fraction whose first full-scene hit belongs to that entity
    must be at least $0.55$ for humans and $0.28$ otherwise.
    Named head meshes, when present, receive an additional test
    with thresholds $0.70$ and $0.35$, respectively.
    Any part of the same entity counts as visible, including for
    head tests; zero-hit tests fail. \\
    Projected readability &
    Frame-required entities tagged as humans, animals, or birds
    must occupy at least $0.20$, $0.13$, or $0.045$ of image
    height, respectively, using clipped projected bounds. \\
    Interaction visibility &
    Contact meshes named as a hand, handle, shaft, or glasses
    bridge require exact-part visibility of at least $0.22$:
    the first scene hit must be that mesh. Nonhuman entities
    participating in \texttt{hold} require entity visibility
    of at least $0.65$. \\
    Relative ordering &
    Left/right relations require a strictly positive signed
    displacement between projected bounding-box centers in the
    requested direction. Front/behind relations use a strictly
    positive camera-depth difference. No additional positive
    margin is imposed; \texttt{above} and \texttt{inside}
    are assessed visually. \\
    Visual requirements &
    Critical checks and every declared relation require
    \texttt{pass}. For other applicable checks, \texttt{fail}
    and \texttt{uncertain} count as failures, while
    \texttt{minor} is tolerated. The critical checks and
    candidate-level aggregation are defined in the text. \\
    \bottomrule
  \end{tabular*}
  \endgroup
\end{table*}

\begin{figure*}[!t]
\centering
\begin{minipage}{\textwidth}
\begin{promptstylebox}{Prompt 3: Data Categorization}
\small

\textbf{Role.}
You categorize training examples by their principal
compositional requirement.

\textbf{Input.}
Reference images, a generation or editing instruction,
and a target image.

\textbf{Task.}
Assign one primary category based on the instruction.
Use the images to resolve visual references, without
treating incidental image content as a requirement.

\textbf{Categories.}
\begin{enumerate}[leftmargin=*,nosep]
  \item \textbf{Shape Alignment.}
  Matching or changing coarse object shape, silhouette,
  or proportions. Excludes appearance-only changes and
  articulated body posture.

  \item \textbf{Object Interaction.}
  Actions and functional relationships between entities,
  including human--human and human--object interactions.
  The main requirement concerns how entities act together,
  or how one uses, manipulates, or responds to another.

  \item \textbf{Spatial Arrangement.}
  Relative placement, depth ordering, distance, or spatial
  containment. The main requirement concerns where
  entities are located relative to one another.

  \item \textbf{Scene Composition.}
  Selecting subjects, specifying instance counts, or
  combining subjects and backgrounds into a scene,
  when no other category is the dominant requirement.

  \item \textbf{Pose Control.}
  Specifying or transferring a subject's body posture or
  articulated configuration. The posture itself is the
  main requirement, rather than an action involving
  another entity.
\end{enumerate}

\textbf{Assignment Rules.}
Identify the main requested operation and distinguish it
from supporting constraints. Choose the category whose
violation would most directly defeat that operation.
Distinguish actions between entities from individual
postures and relative placement. Multiple subjects,
physical proximity, or incidental contact alone do not
establish an interaction.

If categories remain equally central, or none is
supported, return \texttt{null} and set
\texttt{needs\_review} to true. Do not force labels to
match a desired distribution.

\textbf{Output Format.}
Return raw JSON only. Use an exact category name from
the list, or \texttt{null}, and briefly identify the
decisive instruction requirement. The following
illustrates the output format only.

{\ttfamily\footnotesize
\begin{verbatim}
{
  "primary_category": "Pose Control",
  "needs_review": false,
  "reason": "Posture is the main request."
}
\end{verbatim}
}

\end{promptstylebox}
\end{minipage}
\caption{Prompt template for categorizing examples
by their principal compositional requirement.}
\label{fig:prompt_data_categorization}
\end{figure*}

\section{Implementation Details}
\label{sec:supp_implementation}

We implement generator adaptation using DiffSynth-Studio and
Hugging Face Accelerate. For Qwen-Image-Edit-2511 and
FLUX.2-klein-base-9B, we train rank-32 LoRA adapters in the
diffusion transformer while keeping the pretrained parameters
frozen. Both models are trained for five epochs on the 3,187
examples in MAGPIE-Instruct-3K using AdamW with a learning rate
of $10^{-4}$ and weight decay of $0.01$.

Training uses eight H800 GPUs with a per-GPU batch size of one
and no gradient accumulation. For square images, the pixel budget is capped at
1048576 pixels ($1024\times1024$). Gradient checkpointing
is enabled for both generators; Qwen additionally uses
activation offloading during gradient checkpointing.

\section{More Details in the Harness}
\label{sec:supp_verification}

This section details verification and candidate selection within
the MAGPIE harness. A single agent orchestrates scene planning,
construction, verification, and revision. The scene specification
defines entity identifiers and reference associations, required
relations, and framing requirements. Geometric checks use
measurements from the executed Blender meshes and camera, while
visual review uses the agent's assessment of rendered views
against the references and instruction. These checks provide
revision feedback, and the harness programmatically applies the
acceptance and candidate-retention rules to their outputs.

\paragraph{Geometric and visual evidence.}
Table~\ref{tab:operational_verification} specifies the geometric
tests, their thresholds, and their applicability. Entity identifiers
associate measurements with the intended instances and contact
parts. The agent receives the proxy render, localized
relation views, references, instruction, scene specification,
and geometric feedback. It describes the observed composition,
identifies structures to preserve and defects to repair, and
returns \texttt{pass}, \texttt{minor}, \texttt{fail}, or
\texttt{uncertain} judgments with visual evidence. A valid review
must follow this structured schema and provide nonempty evidence.
Contact assessment combines the distance test with visual assessment
of the requested pose and interaction.

\paragraph{Visual decision rules.}
Critical checks include entity inventory, reference binding,
spatial layout, pose/contact, and source preservation for editing.
These checks and every declared relation must receive \texttt{pass};
a \texttt{minor}, \texttt{fail}, or \texttt{uncertain} label on any
of them counts as a visual failure. Source preservation is
inapplicable to generation and treated as satisfied. For other
checks and applicable requirements, \texttt{fail} and
\texttt{uncertain} count as failures, while \texttt{minor} is
tolerated. Appearance-only requirements may be inapplicable to
a structural proxy, and its neutral colors and materials are allowed.

\paragraph{Acceptance and candidate retention.}
For a successfully rendered candidate $c^{(k)}$ with a valid
visual review, let $G_k$ and $V_k$ count failed geometric and
visual diagnostic records, respectively. The channels are
recorded separately: a defect detected by both contributes
to both counts. Let $M_k$ count \texttt{minor} labels among
fixed visual checks and applicable requirements; relation
labels enter $V_k$. The score and retained candidate are
\begin{equation}
\begin{aligned}
  s_k &= (G_k+V_k,\;G_k,\;M_k,\;k),\\
  k^\star &=
  \operatorname*{arg\,min}_{k\in\mathcal{K}}^{\mathrm{lex}}\;s_k,
\end{aligned}
\label{eq:explicit_candidate_selection}
\end{equation}
where $\mathcal{K}$ contains the iteration indices of valid
reviewed candidates; the retained code and render correspond
to $k^\star$. Lexicographic ordering prioritizes fewer total failures,
then fewer geometric failures, then fewer minor judgments.
If these three counts tie, the earlier candidate is retained.
Geometric failures therefore have priority when total failure
counts tie. Runtime acceptance requires $G_k=V_k=0$, so neither
a visual pass nor a geometric pass alone establishes acceptance.
A tolerated minor judgment does not prevent it.

\paragraph{Revision and stopping.}
Each revision starts from the retained code and render with
their corresponding feedback. Feedback from a rejected
proposal remains associated with that proposal. The agent
is instructed to preserve satisfied relations while repairing
identified defects; each revised program is executed,
rendered, and reviewed again. Under the default budget,
the harness performs at most six refinement rounds and
stops early when $G_k=V_k=0$. At the budget limit, it
returns the retained candidate, marked \texttt{needs\_review}
if failures remain. Invalid submissions, failed renders,
and malformed reviews do not replace the retained candidate;
if the budget is exhausted without a valid reviewed render,
construction returns an error. Diagnostics, visual evidence,
selection scores, and retained-candidate identifiers are logged
together, linking each retention decision to its supporting evidence.

\section{Additional Evaluation Analysis}
\label{sec:additional_evaluation_analysis}

We analyze the computation required for scene construction,
compare against additional sampling under comparable time budgets,
and examine how refinement affects generation quality.
We then assess the structural quality of the selected proxies
through human evaluation.

\subsection{Inference Time Analysis}
\label{sec:inference_efficiency}

\begin{table*}[!t]
  \caption{\textbf{Inference time breakdown.}
  Mean recorded stage time per example in seconds.
  Plan, Init., Refine, and Synth. denote scene planning,
  initial proxy construction, verification and refinement,
  and image synthesis, respectively. Total is the sum of
  the displayed stage means, excluding untimed overhead.
  Qwen and FLUX denote Qwen-Image-Edit-2511 and
  FLUX.2-klein-base-9B. A dash denotes an absent stage.}
  \label{tab:supp_runtime}
  \centering
  \begingroup
  \rmfamily
  \small
  \setlength{\tabcolsep}{3pt}
  \renewcommand{\arraystretch}{1.10}
  \begin{tabular*}{\textwidth}{@{}ll@{\extracolsep{\fill}}ccccc@{}}
    \toprule
    \textbf{Benchmark} & \textbf{Method / Setting}
    & \textbf{Plan} & \textbf{Init.} & \textbf{Refine}
    & \textbf{Synth.} & \textbf{Total} $\downarrow$ \\
    \midrule
    \multirow{4}{*}{DreamOmni2Bench}
    & Qwen & -- & -- & -- & 77.20 & 77.20 \\
    & \quad + MAGPIE & 20.65 & 56.32 & 195.20 & 103.01 & 375.18 \\
    & FLUX & -- & -- & -- & 60.97 & 60.97 \\
    & \quad + MAGPIE & 20.65 & 56.32 & 195.20 & 84.55 & 356.72 \\
    \midrule
    \multirow{4}{*}{SconeEval}
    & Qwen & -- & -- & -- & 78.50 & 78.50 \\
    & \quad + MAGPIE & 27.33 & 85.18 & 224.72 & 107.00 & 444.23 \\
    & FLUX & -- & -- & -- & 69.76 & 69.76 \\
    & \quad + MAGPIE & 27.33 & 85.18 & 224.72 & 78.54 & 415.77 \\
    \bottomrule
  \end{tabular*}
  \endgroup
\end{table*}

\paragraph{Runtime breakdown.}
Table~\ref{tab:supp_runtime} decomposes recorded inference time
into scene planning, initial proxy construction, verification and
refinement, and image synthesis. Planning measures scene-planning
model calls. Initial construction includes code generation and
Blender execution and rendering up to the first successful proxy,
including preceding unsuccessful attempts. Refinement includes
proxy verification, subsequent code revisions, and the corresponding
Blender execution and re-rendering. Each recorded operation is
assigned to exactly one stage. Uninstrumented input processing,
validation outside these calls, and orchestration are excluded.

The timing analysis aggregates recorded model calls and
Blender executions within the first six construction cycles,
including timed retries, failed attempts, and timeouts;
subsequent cycles are excluded. Construction statistics
cover all 319 DreamOmni2Bench examples and 409 SconeEval
examples, all of which produced successful renders.
Synthesis statistics include 319 outputs for both the
baseline and MAGPIE on DreamOmni2Bench, and 409 outputs
for each synthesis setting on SconeEval.

Proxy construction takes 272.17\,s per example on DreamOmni2Bench
and 337.23\,s on SconeEval. Verification and refinement account
for the largest share, at 195.20\,s and 224.72\,s, respectively.
Blender execution and rendering contribute 32.35\,s and 38.10\,s,
corresponding to 11.9\% and 11.3\% of construction time.
Including synthesis, the recorded stage sums are 375.18\,s and
444.23\,s for Qwen, and 356.72\,s and 415.77\,s for FLUX.
The corresponding baseline synthesis times are 77.20\,s and
78.50\,s for Qwen, and 60.97\,s and 69.76\,s for FLUX.
These measurements identify iterative construction as the main
source of additional inference cost.

\begin{table}[!t]
  \caption{\textbf{Comparison under comparable inference time budgets on SconeEval.}
  Budgeted base uses best-of-six sampling with manual selection.
  Scores range from 0 to 10; bold marks the best score within
  each backbone group. Human selection time is excluded.}
  \label{tab:supp_matched_time}
  \centering
  \small
  \setlength{\tabcolsep}{3pt}
  \renewcommand{\arraystretch}{1.10}
  \begin{tabular*}{\linewidth}{@{\extracolsep{\fill}}lcccc@{}}
    \toprule
    \textbf{Method}
    & \textbf{COM} $\uparrow$
    & \textbf{DIS} $\uparrow$
    & \textbf{Overall} $\uparrow$ \\
    \midrule
    Qwen-Image-Edit-2511
    & 8.21 & 8.23 & 8.22 \\
    \quad Budgeted base
    & 8.59 & 8.50 & 8.54 \\
    \quad + MAGPIE
    & \textbf{8.60} & \textbf{9.26} & \textbf{8.93} \\
    \midrule
    FLUX.2-klein-base-9B
    & 8.20 & 7.93 & 8.07 \\
    \quad Budgeted base
    & 8.48 & 8.36 & 8.42 \\
    \quad + MAGPIE
    & \textbf{8.68} & \textbf{9.43} & \textbf{9.05} \\
    \bottomrule
  \end{tabular*}
\end{table}

\paragraph{Comparison under comparable time budgets.}
To examine whether additional sampling alone accounts for the
gains, we equip each base generator with a best-of-six strategy
on SconeEval. For each example, we sequentially generate six
candidates using different random seeds and manually select the
best image according to the instruction and references. The selected
image is evaluated using the standard SconeEval protocol.

Using the mean baseline synthesis times in
Table~\ref{tab:supp_runtime}, six sequential runs correspond to
approximately 471.00\,s for Qwen and 418.56\,s for FLUX,
compared with MAGPIE's recorded stage sums of 444.23\,s and
415.77\,s. These budgets are comparable rather than exactly
matched, and the estimates exclude untimed overhead.
Table~\ref{tab:supp_matched_time} shows that best-of-six sampling
improves both backbones over single-sample generation, but MAGPIE
achieves higher overall scores: 8.93 versus 8.54 for Qwen and
9.05 versus 8.42 for FLUX. Thus, on these evaluated settings, comparable additional sampling does not recover the full benefit of proxy-guided generation.

\subsection{Effect of Refinement Budget}
\label{sec:refinement_budget}

\begin{table}[!t]
\caption{\textbf{Refinement budget, performance, and construction time.}
The adapted FLUX generator is fixed across all settings.
Edit., Gen., and Overall are DreamOmni2Bench success
rates (\%). Proxy time includes scene planning, initial
proxy construction, and refinement. Best scores are highlighted in bold.}
  \label{tab:supp_refinement_budget}
  \centering
  \begingroup
  \rmfamily
  \small
  \setlength{\tabcolsep}{3pt}
  \renewcommand{\arraystretch}{1.10}
  \begin{tabular*}{\columnwidth}{@{}c@{\extracolsep{\fill}}cccc@{}}
    \toprule
    \textbf{Max. rounds}
    & \textbf{Edit.} $\uparrow$
    & \textbf{Gen.} $\uparrow$
    & \textbf{Overall} $\uparrow$
    & \textbf{Proxy Time (s)} $\downarrow$ \\
    \midrule
    0 & 54.63 & 51.75 & 53.61 & 76.97 \\
    3 & 60.98 & 61.40 & 61.13 & 211.44 \\
    6 & \textbf{66.83} & \textbf{67.54}
      & \textbf{67.08} & 272.17 \\
    \bottomrule
  \end{tabular*}
  \endgroup
\end{table}

\begin{figure}[!t]
  \centering
  \includegraphics[width=\linewidth]{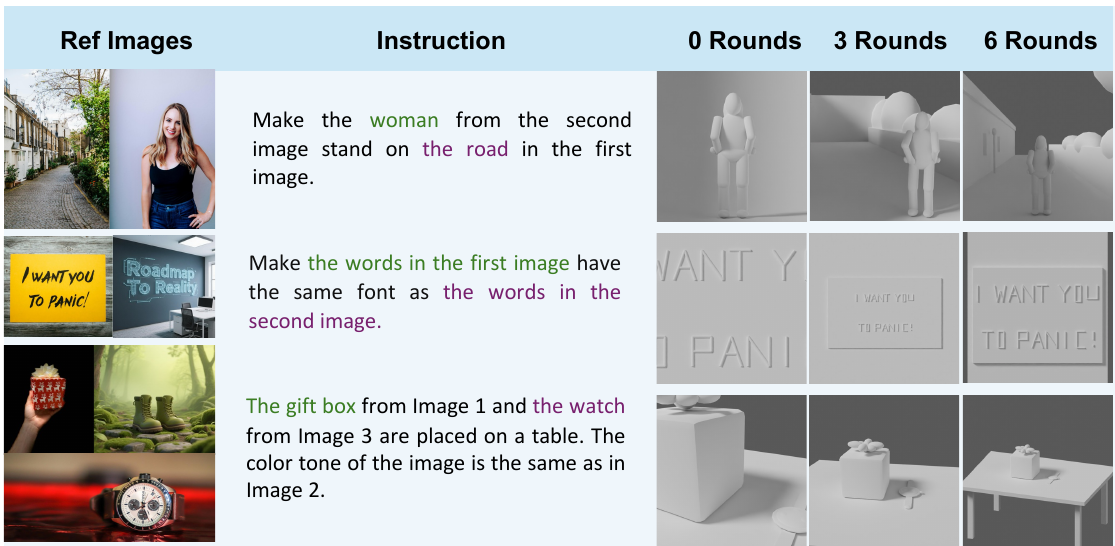}
  \caption{\textbf{Proxy scenes across refinement budgets.}
  Each row shows the reference images, instruction, and proxies
  selected with maximum budgets of 0, 3, and 6 refinement rounds.}
  \label{fig:supp_refinement_evolution}
\end{figure}

\begin{table*}[!t]
  \caption{\textbf{Human evaluation of proxy quality on
  DreamOmni2Bench.}
  Each selected proxy receives one overall quality label.
  $N$ denotes the number of audited examples.
  Percentages are computed within each task group.
  Cohen's $\kappa$ measures annotator agreement before
  disagreement resolution. A dash denotes an unavailable
  agreement statistic.}
  \label{tab:supp_reliability}
  \centering
  \begingroup
  \rmfamily
  \small
  \setlength{\tabcolsep}{3pt}
  \renewcommand{\arraystretch}{1.10}
  \begin{tabular*}{\textwidth}
    {@{}l@{\extracolsep{\fill}}ccccc@{}}
    \toprule
    \textbf{Task group}
    & \textbf{$N$}
    & \textbf{Correct (\%)}
    & \textbf{Minor issue (\%)}
    & \textbf{Major error (\%)}
    & \textbf{Cohen's $\kappa$} \\
    \midrule
    Abstract editing
    & 164 & 92.07 & 7.93 & 0.00 & 0.827 \\
    Concrete editing
    & 41 & 95.12 & 4.88 & 0.00 & 0.787 \\
    Abstract generation
    & 90 & 86.67 & 11.11 & 2.22 & 0.854 \\
    Concrete generation
    & 24 & 100.00 & 0.00 & 0.00 & - \\
    \bottomrule
  \end{tabular*}
  \endgroup
\end{table*}

\paragraph{Quality--time trade-off.}
We compare maximum budgets of 0, 3, and 6 refinement rounds
on DreamOmni2Bench using the same adapted FLUX generator.
Zero rounds uses the initial proxy before render-guided refinement.
The harness stops early when all verification criteria are satisfied
and retains the best candidate within each budget using
Eq.~\ref{eq:explicit_candidate_selection}.
Table~\ref{tab:supp_refinement_budget} reports success rates and
proxy construction time, including planning, initial construction,
and refinement but excluding final image synthesis.

Both editing and generation success improve across the evaluated
budgets. Overall success rises from 53.61\% at zero rounds to
61.13\% at three rounds and 67.08\% at six rounds, while
construction time increases from 76.97\,s to 211.44\,s and
272.17\,s. Extending the budget from three to six rounds
therefore yields a further 5.95 percentage-point improvement
at an additional construction cost of 60.73\,s. Further
refinement improves task success, with additional computation
required for verification and revision.

\paragraph{Qualitative evolution.}
Figure~\ref{fig:supp_refinement_evolution} compares the proxies
selected at each budget for the same inputs. In the first example,
the isolated figure is revised into a scene that includes the road
and surrounding geometry. In the text example, refinement changes
the lettering from sparse, incomplete geometry into a fuller
representation of the requested phrase. In the final example,
the gift box and watch are reorganized on a table, making their
relative placement more explicit. These examples illustrate how program revisions make the requested
structure more explicit before image synthesis.

\subsection{Construction Reliability}
\label{sec:construction_reliability}
\label{sec:human_proxy_evaluation}

\paragraph{Human evaluation of proxy quality.}
Two annotators independently reviewed 319 final proxy renderings
returned by the harness (Sec.~\ref{sec:supp_verification}) on
DreamOmni2Bench, covering 164
abstract editing, 41 concrete editing, 90 abstract generation,
and 24 concrete generation examples. They inspected the references,
instruction, and selected proxy for entity inclusion and counts,
spatial arrangements, poses and interactions, and framing and
visibility. Appearance differences inherent to simplified proxy
geometry were excluded.

Each proxy receives one overall label: \emph{correct}, when the
requested composition is satisfied; \emph{minor issue}, when a
localized discrepancy does not substantially alter it; or
\emph{major error}, when an essential structural requirement
is violated. Table~\ref{tab:supp_reliability} reports the final
label distributions within each task group and Cohen's $\kappa$
computed from independent annotations before disagreement resolution.

Across the audited examples, 292 (91.54\%) are correct,
25 (7.84\%) contain minor issues, and two (0.63\%) contain
major errors. Abstract generation has the lowest correctness rate
(86.67\%) and accounts for both major errors. The other groups
have correctness rates of 92.07\%--100.00\%, with no major
errors observed. Available inter-annotator agreement values range
from $\kappa=0.787$ to $0.854$. These results indicate that
most selected proxies preserve the requested composition, while
abstract generation remains the most error-prone audited group.

\begin{figure*}[!t]
  \centering
  \includegraphics[width=\linewidth]{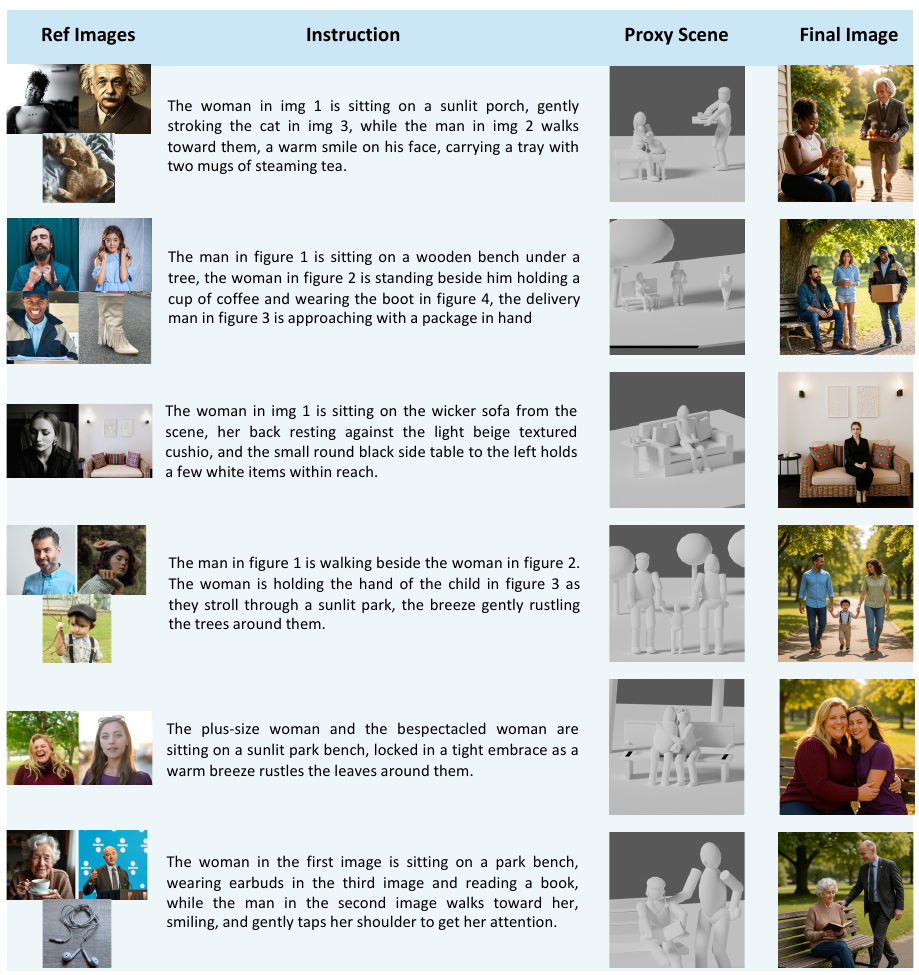}
  \caption{\textbf{Qualitative results on SconeEval.}
  Each row shows the reference images, instruction, selected proxy,
  and final image generated with MAGPIE. The examples cover
  multi-subject arrangements, articulated poses, and interactions.}
  \label{fig:supp_sconeeval_examples}
\end{figure*}

\begin{figure*}[!t]
  \centering
  \includegraphics[width=\linewidth]{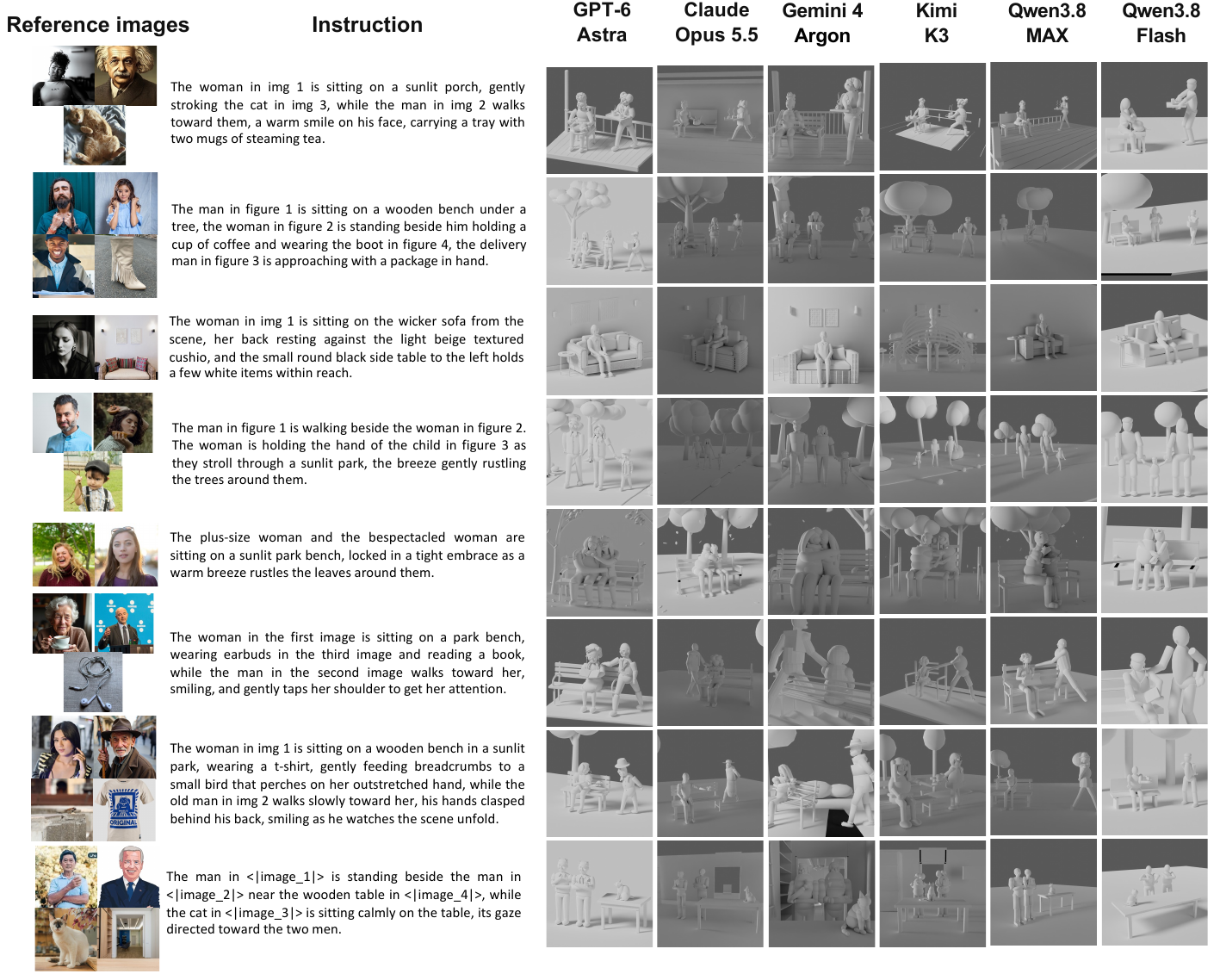}
  \caption{\textbf{Proxy scene comparison across coding agents.}
  Each group shares the same reference images and instruction
  and compares the proxy renderings returned by six agents
  using the same MAGPIE harness. The examples compare subject inclusion, spatial arrangement,
  and the geometric realization of poses and interactions.}
  \label{fig:supp_agent_comparison}
\end{figure*}

\section{Additional Qualitative Results}
\label{sec:more_experimental_results}

\subsection{Proxy-Guided Generation on SconeEval}
\label{sec:supp_sconeeval}

Figure~\ref{fig:supp_sconeeval_examples} pairs selected proxies
with the final images generated by MAGPIE. The examples combine
multiple reference subjects with requirements involving seating,
walking, hand holding, and other interactions. The proxies make
subject placement and coarse body configurations explicit, while
the synthesized images recover appearance from the references.
These examples complement the aggregate SconeEval scores by
showing the intermediate structural guidance alongside its final
image realization.

\subsection{Comparison across Coding Agents}
\label{sec:agent_comparison}

\paragraph{Evaluation setup.}
We qualitatively examine how coding agents affect scene
construction within MAGPIE. We compare GPT-6 Astra,
Claude Opus 5.5, Gemini 4 Argon, Kimi K3, Qwen3.8-Max,
and Qwen3.8-Flash. All agents receive the same references and
instruction and share the specification format, procedural
modeling interfaces, geometric checks, and refinement rules.
We inspect subject inclusion and counts, spatial arrangements,
poses, and interactions.

\paragraph{Qualitative comparison.}
Figure~\ref{fig:supp_agent_comparison} shows variation in geometric
detail and articulation across agents. For example, GPT-6 Astra
constructs detailed human and furniture geometry in the porch
and sofa scenes, while Qwen3.8-Flash uses simpler geometry that
retains the main arrangements and interactions, including
embracing and shoulder tapping. The examples illustrate how
the same harness accommodates different construction models.

\clearpage
\twocolumn

\end{document}